\documentclass[10pt,twocolumn,letterpaper]{article}

\usepackage{cvpr}
\usepackage{graphicx, amsmath, amssymb, booktabs}
\usepackage[pagebackref,breaklinks,colorlinks]{hyperref}
\usepackage[capitalize]{cleveref}
\usepackage[accsupp]{axessibility}
\crefname{section}{Sec.}{Secs.}
\Crefname{section}{Section}{Sections}
\Crefname{table}{Table}{Tables}
\crefname{table}{Tab.}{Tabs.}

\usepackage{enumitem, graphicx, multirow, amsfonts, amsmath, xcolor, verbatim, algorithm, algorithmicx, algpseudocode, booktabs, arydshln, times, epsfig, amssymb, pifont, capt-of, wrapfig, lipsum}

\def\cvprPaperID{11551}
\def\confName{CVPR}
\def\confYear{2022}

\begin{document}

\title{M$^3$L: Language-based Video Editing via Multi-Modal Multi-Level Transformers}
\author{Tsu-Jui Fu$^\dagger$, Xin Eric Wang$^\ddagger$, Scott T. Grafton$^\dagger$, Miguel P. Eckstein$^\dagger$, William Yang Wang$^\dagger$ \\
\\
$^\dagger$UC Santa Barbara~~$^\ddagger$UC Santa Cruz \\
{\tt \small tsu-juifu@ucsb.edu, \{scott.grafton, miguel.eckstein\}@psych.ucsb.edu} \\
{\tt \small william@cs.ucsb.edu, xwang366@ucsc.edu}
}
\maketitle

\begin{abstract}
Video editing tools are widely used nowadays for digital design. Although the demand for these tools is high, the prior knowledge required makes it difficult for novices to get started. Systems that could follow natural language instructions to perform automatic editing would significantly improve accessibility. This paper introduces the language-based video editing (LBVE) task, which allows the model to edit, guided by text instruction, a source video into a target video. LBVE contains two features: 1) the scenario of the source video is preserved instead of generating a completely different video; 2) the semantic is presented differently in the target video, and all changes are controlled by the given instruction. We propose a Multi-Modal Multi-Level Transformer (M$^3$L) to carry out LBVE. M$^3$L dynamically learns the correspondence between video perception and language semantic at different levels, which benefits both the video understanding and video frame synthesis. We build three new datasets for evaluation, including two diagnostic and one from natural videos with human-labeled text. Extensive experimental results show that M$^3$L is effective for video editing and that LBVE can lead to a new field toward vision-and-language research.
\end{abstract}

\section{Introduction}
Video is one of the most direct ways to convey information, as people are used to interacting with this world via dynamic visual perception. Nowadays, video editing tools like Premiere and Final Cut are widely applied for digital design usages, such as film editing or video effects. 
However, those applications require prior knowledge and complex operations to utilize successfully, which makes it difficult for novices to get started. For humans, natural language is the most natural way of communication. If a system can follow the given language instructions and automatically perform related editing actions, it will significantly improve accessibility and meet the considerable demand. 

\begin{figure}[t]
\centering
    \includegraphics[width=\linewidth]{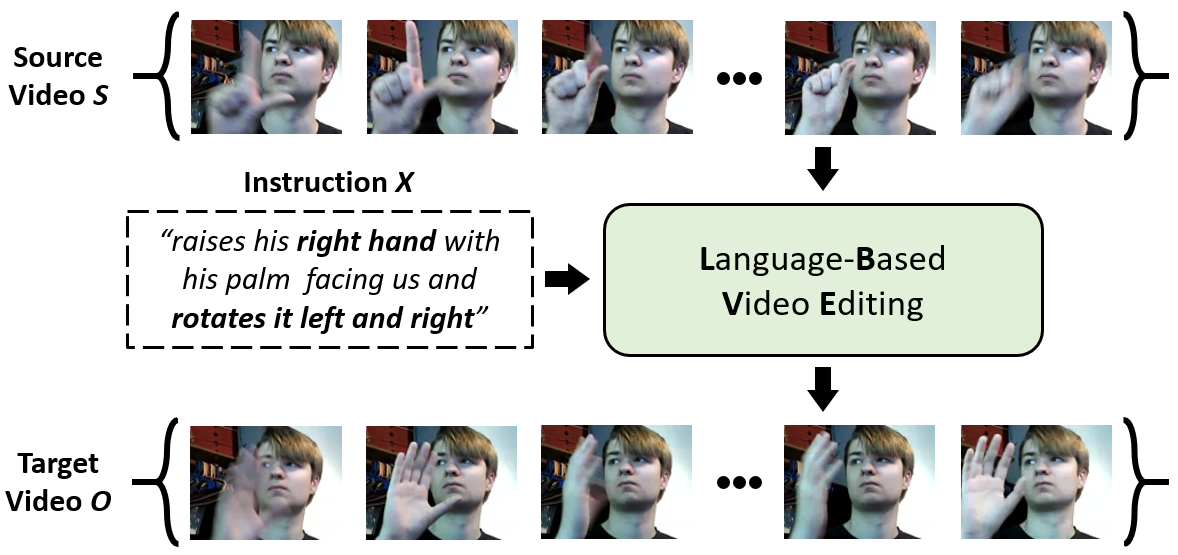}
    \vspace{-3ex}
    \caption{The introduced language-based video editing (LBVE) task. LBVE requires to edit a source video $S$ into the target video $T$ guided by the instruction $X$.}
    \vspace{-3ex}
    \label{fig:intro}
\end{figure}

In this paper, we introduce language-based video editing (LBVE), a general V2V task, where the target video is controllable directly by language instruction. LBVE treats a video and an instruction as the input, and the target video is edited from the textual description. As illustrated in Fig.~\ref{fig:intro}, the same person performs different hand gestures guided by the instruction. Different from text-to-video (T2V) \cite{li2018t2v,balaji2019t2v,pan2017t2v,marwah2017c2v}, video editing enjoys two following feature: 1) the scenario (e.g., scene or humans) of the source video is preserved instead of generating all content from scratch; 2) the semantic (e.g., property of the object or moving action) is presented differently in the target video. The main challenge of LBVE is to link the video perception with language understanding and reflect what semantics should be manipulated during the video generation but under a similar scenario. People usually take further editing steps onto a base video rather than create all content from the beginning. We believe that our LBVE is more practical and corresponding to human daily usage. 

\begin{figure*}[t]
\centering
    \includegraphics[width=\linewidth]{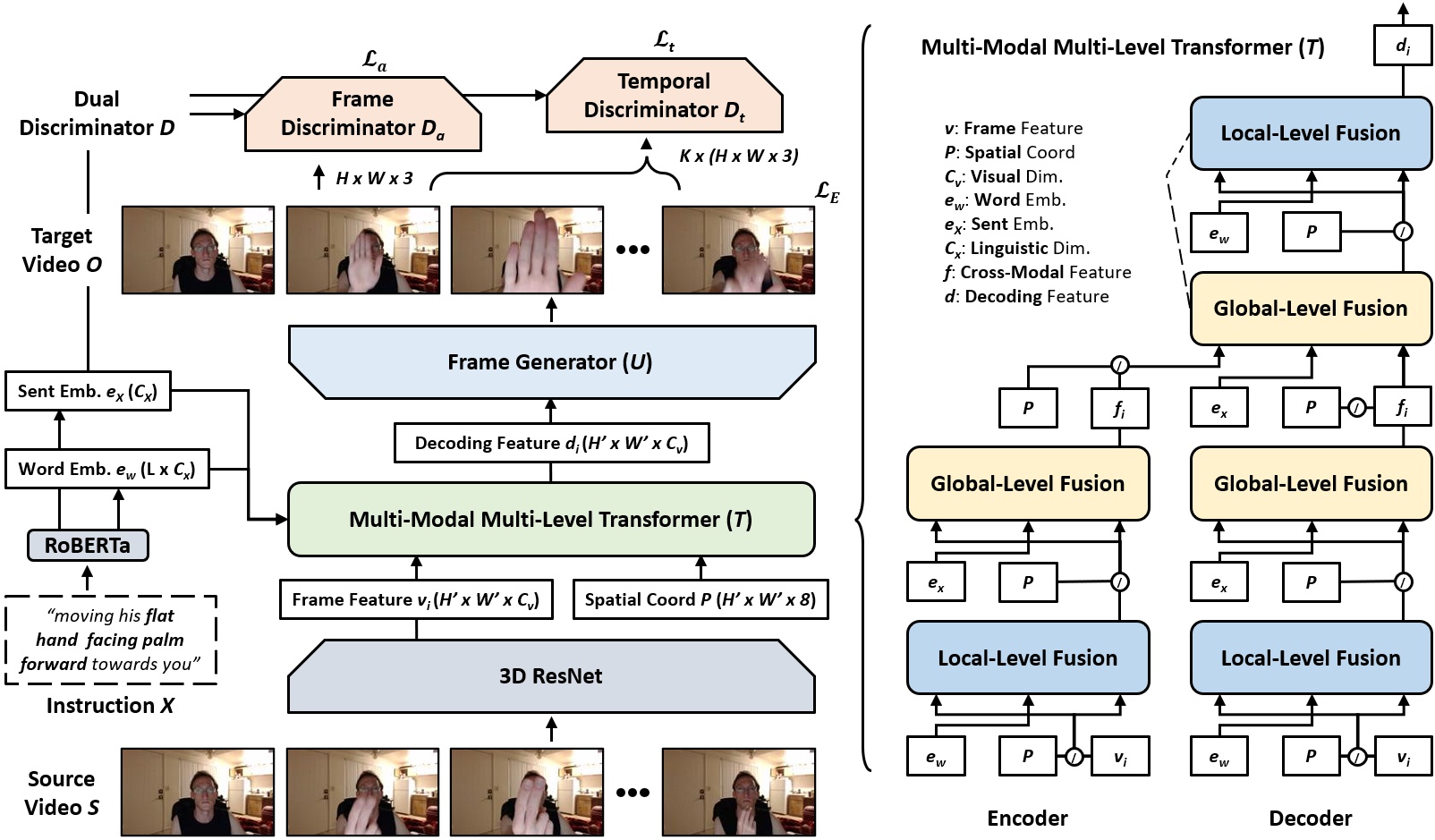}
    \vspace{-3ex}
    \caption{An overview architecture of our multi-modal multi-level transformer (M$^3$L). M$^3$L contains the multi-modal multi-level transformer $T$ to encode the source video $S$ and decode for the target video frame $o$ by the multi-level fusion (MLF).}
    \vspace{-3ex}
    \label{fig:overview}
\end{figure*}

To tackle the LBVE task, we propose a multi-modal multi-level transformer (M$^3$L) to perform video editing conditioning on the guided text. As shown in Fig.~\ref{fig:overview}, M$^3$L contains a multi-modal multi-level Transformer where the encoder models the moving motion to understand the entire video, and the decoder serves as a global planner to generate each frame of the target video. For better video perception to link with the given instruction, the incorporated multi-level fusion fuses between these two modalities. During encoding, the local-level fusion is applied with the text tokens for fine-grained visual understanding, and the global-level fusion extracts the key feature of the moving motion. Reversely, during decoding, we first adopt global-level fusion from whole instruction to give a high-level plan for the target video, and then the local-level fusion can further generate each frame in detail with the specific property. With multi-level fusion, M$^3$L learns explicit vision-and-language perception between the video and given instruction, yielding better video synthesis.

For evaluation, we collect three datasets under the brand-new LBVE task. There are E-MNIST and E-CLEVR, where we build from hand-written number recognition MNIST \cite{lecun2010mnist} and compositional VQA CLEVR \cite{johnson2017clevr}, respectively. Both E-MNIST and E-CLEVR are prepared for evaluating the content replacing (different numbers or shapes and colors) and semantic manipulation (different moving directions or related positions). As a new task, diagnostic datasets help analyze the progress and discover the shortcomings. To investigate the capability of LBVE for natural video with open text, E-JESTER is built upon the same person performing different hand gestures with human instruction.

Our experimental results show that the multi-modal multi-level transformer (M$^3$L) can carry out the LBVE task, and the multi-level fusion further helps between video perception and language understanding in both aspects of content replacing and semantic manipulation. In summary, our contributions are four-fold:
\begin{itemize}[noitemsep, leftmargin=*]
    \item We introduce the LBVE task to manipulate video content controlled by text instructions.
    \item We present M$^3$L to perform LBVE, where the multi-level fusion further helps between video perception and language understanding.
    \item For evaluation under LBVE, we prepare three new datasets containing two diagnostic and one natural video with human-labeled text.
    \item Extensive ablation studies show that our M$^3$L is adequate for video editing, and LBVE can lead to a new field toward vision-and-language research.
\end{itemize}

\section{Related Work}
\vspace{0.5ex} \noindent\textbf{Language-based Image Editing.}
Different from text-to-image (T2I) \cite{nguyen2017langevin,reed2017pixcelcnn, tan2019t2s}, which generates an image that matches the given instruction, language-based image editing (LBIE) understands the visual difference and edits between two images based on the guided text description. Image Spirit \cite{cheng2013image-spirit} and PixelTone \cite{laput2013pixel-tone} first propose the LBIE framework but accept only rule-based instruction and pre-defined semantic labels, which limits the practicality of LBIE. Inspired by numerous GAN-based methods \cite{reed2016t2i-gan,zhang2017stack-gan, xu2018att-gan} in T2I, there are some previous works \cite{chen2018lbie-ram,shinagawa2017edit-mnist} perform LBIE as image colorization by the conditional GAN. Since humans do not always finish editing all-at-once but will involve several different steps, iterative LBIE (ILBIE) \cite{elnouby2018ilbie,fu2020sscr} is proposed to imitate the actual process by the multi-turn manipulation and modeling the instructed editing history. Similar to LBIE, language-based video editing (LBVE) is to edit the content in a video by the guided instruction. To perform LBVE, it is required to model the dynamic visual perception instead of just a still image and consider the temporal consistency of each frame during the generation to make a smooth result video.

\vspace{0.5ex} \noindent\textbf{Language-based Video Generation.}
Generative video modeling \cite{villegas2017dec-motion,brabandere2016dfn,balaji2019t2v,pu2018v2t,liu2017vido-synthesis,srivastava2015m-mnist,kalchbrenner2017vpn,weissenborn2020scale-avm,babaeizadeh2017svp,denton2018svg,hsieh2018ddpae,saito2018tgan,saito2017tgan,vondrick2016tiny-video,mathieu2016adv+gdl,tulyakov2017mocogan,clark2019dvd-gan,hu2018vs-stroke,hao2018vg-traj,marwah2017c2v,fried2019talking-head} is a widely-discussed research topic that looks into the capability of a model to generate a video purely in pixel space. Built upon video generation, text-to-video (T2V) \cite{li2018t2v,balaji2019t2v,pan2017t2v,marwah2017c2v} synthesizes a video by the guided text description, which makes the video output controllable by the natural language. In this paper, we investigate the video editing task, which replaces the specific object with different properties or changes the moving motion in the input video. Different from generating video from scratch, video editing requires extracting the dynamic visual perception of the source video and manipulating the semantic inside to generate the target video.

\vspace{0.5ex} \noindent\textbf{Video-to-Video Synthesis.}
Video super-resolution \cite{jo2018vsr,liu2017vsr}, segmentation video reconstruction \cite{wang2018vid2vid,wang2019vid2vid}, video style transfer \cite{chen17vst,xia21vst,deng21vst}, or video inpainting \cite{chang19vornet,kim2019dvi,xu2019dvi} can be considered as the particular case of video-to-video synthesis (V2V). Since they all depend on the task themselves, the variability between source-target is still under the problem-specific constraint. Among them, video prediction \cite{reda2018sdcnet,walker2021vp,li2020vp,guen2020vp}, which predicts future frames conditioning on the given video, is one of the most related to our present LBVE task. 
Both video prediction and LBVE should understand the hidden semantic of the given video first and then predict the target frames with different content inside. While for video prediction, there are many possibilities of appeared future events, which makes it not deterministic for real-world usage \cite{marwah2017c2v}. On the other hand, LBVE is controllable by the given instruction, which involves both content replacing (object changing) and semantic manipulation (moving action changing). With the guided text description, LBVE can perform V2V with content editing and lead to predictable target video.

\section{Language-based Video Editing}
\subsection{Task Definition}
We study the language-based video editing (LBVE) task to edit a source video $S$ into a target video $O$ by a given instruction $X$, as shown in Fig.~\ref{fig:intro}. Specifically, the source video $S$ contains $N$ frames as $\{s_1, s_2, ..., s_N\}$, and the instruction $X=\{w_1, w_2, ..., w_L\}$ where $L$ is the number of word token in $X$. The target video $O$ also includes $N$ frames as $\{o_1, o_2, ..., o_N\}$.
For LBVE, the model should preserve the scenario from $S$ but change the related semantics in $O$ guided by $X$. Note that the editing process is at a pixel level where the model has to generate each pixel of each frame and then assemble them as the target video.

\subsection{Overview}
An overview of our multi-modal multi-level transformer (M$^3$L) for LBVE is illustrated in Fig.~\ref{fig:overview}. 
M$^3$L first extracts the frame feature $v_i$ for the frame $s_i$ in the source video $S$; the sentence embedding $e_X$ and each word embedding $e_w$ for the instruction $X$. Then, the multi-modal multi-level transformer $T$ is proposed to model the sequential information of the source and the target video as the decoding feature $d_i$. In particular, the multi-level fusion (MLF) performs the cross-modal fusion between video $v$ and instruction $\{e_X, e_w\}$. The local-level fusion (LF) extracts which portion is perceived by token $e_w$ across all words in $X$. Besides, the global-level fusion (GF) models the interaction between the entire video perception and the semantic motion from the whole instruction $e_X$. Finally, with $d_i$, the generator $U$ generates the frame $o_i$ in the target video $O$. In addition, we apply the dual discriminator $D$, where the frame discriminator $D_a$ helps the quality of every single frame, and the temporal discriminator $D_t$ maintains the consistency as a smooth output video.

\vspace{0.5ex} \noindent\textbf{Frame and Linguistic Feature Extraction.}
To perform the LBVE task, We first apply 3D ResNet and RoBERTa \cite{liu2019roberta} to extract the frame feature $v$ and linguistic feature $\{e_X, e_w\}$ for the two modalities independently:
\begin{equation} \small
\begin{split}
    \{v_1, v_2, ..., v_N\} &= \text{3D ResNet}(\{s_1, s_2, ..., s_N\}), \\
    e_X, \{e_{w_1}, e_{w_2}, ..., e_{w_L}\} &= \text{RoBERTa}(X),
\end{split}
\end{equation}
where $e_{w_i}$ is the word embedding of each token $w_i$, $e_X$ is the entire sentence embedding of $X$, and $L$ represents the length of the instruction $X$. In detail, $v \in \mathbb{R}^{H' \times W' \times C_v}$ and each $e \in \mathbb{R}^{C_x}$, where $C_v$ and $C_x$ is the feature dimension of vision and language, respectively. 

\subsection{Multi-Modal Multi-Level Transformer}
As illustrated in Fig.~\ref{fig:overview}, with the frame feature $v$ and linguistic feature $\{e_X, e_w\}$ as the inputs, the multi-modal multi-level transformer $T$ contains an encoder to model the sequential information of the source video $S$ with the given instruction $X$, and a decoder to acquire the decoding feature $d_i$ for generating the target video frame $o_i$. Both the encoder and decoder are composed of multi-level fusion (MLF), which is applied to fuse between vision and language with aspects from different levels.

\begin{figure}[t]
\centering
    \includegraphics[width=\linewidth]{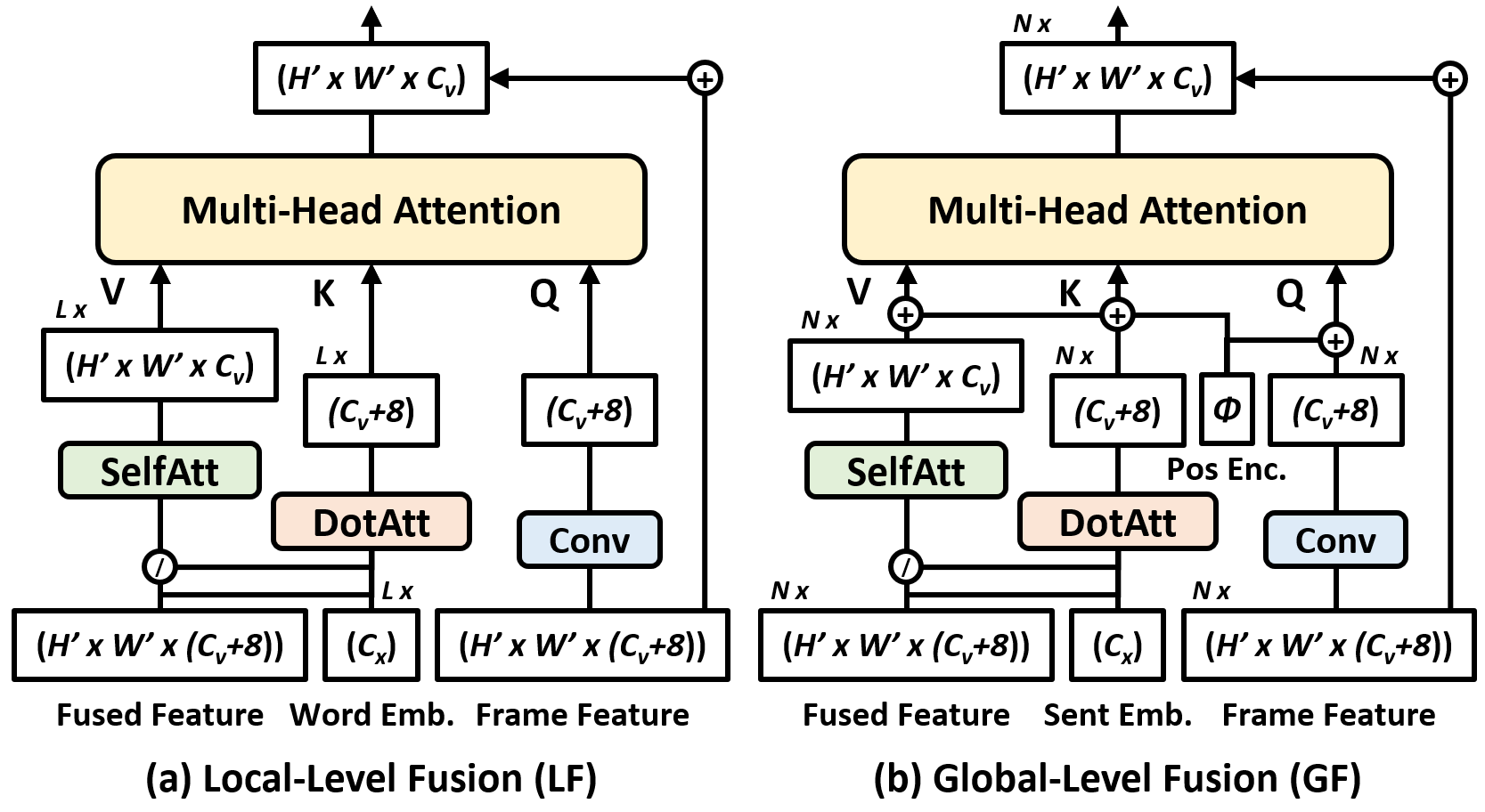}
    \vspace{-3ex}
    \caption{The computing flow of multi-level fusion (MLF), including local-level fusion (LF) and global-level fusion (GF). }
    \vspace{-3ex}
    \label{fig:mlf}
\end{figure}

\vspace{0.5ex} \noindent\textbf{Multi-Level Fusion}
Both video and language are multi-level conveyed, where video is composed of a series of image frames and language is a set of word tokens with a specific order. The multi-level fusion (MLF) consists of the local-level fusion (LF) to fuse between a single frame and each word token, and the global-level fusion (GF) models the entire video sequence with the whole instruction. The computation flow of MLF is illustrated in Fig.~\ref{fig:mlf}. Both LF and GF are computed with the multi-head attention (MHA) \cite{vaswani2017att-all}. MHA acquires the weighted-sum of the value feature (V) by considering the correlation between the query feature (Q) and the key feature (K):
\begin{equation} \small
    \text{MHA}(\text{Q}, \text{K}, \text{V}) = \text{softmax}(\frac{\text{Q} \cdot \text{K}^T}{\sqrt{C_\text{K}}})\text{V}.
\end{equation}

For the local-level fusion (LF), it investigates which portion should be focused by each word $e_w$ in a single frame $v_i$. We provide the relative spatial information by concatenating a 8-D spatial coordinate feature $P$ \cite{liu2017rmi} with $v_i$ as $\mathrm{p}^\text{L}$ . To fuse between vision and language, we apply the self-attention mechanism (SelfAtt) \cite{zhang2019sa-gan,ye2019cmsa} upon the concatenated feature $\mathrm{q}^\text{L}$ to capture the correlation between word expression and visual context into $\mathrm{s}^\text{L}$. Different from CMSA \cite{ye2019cmsa}, which concatenates frame feature with all token embedding directly, our LF further considers the importance of each token. We adopt a 1-layer convolutional net (Conv) to extract the context-only visual feature $\mathrm{c}^\text{L}$ along the channel of $v_i$; and the widely-used dot-product attention (DotAtt) \cite{xu2015visual-att,chaplot2018gated-att} for the word-focused visual feature $\mathrm{d}^\text{L}_l$ with each word $e_{w_l}$. Therefore, the correlation between $\mathrm{c}^\text{L}$ and $\mathrm{d}^\text{L}_l$ can be considered as the important portion of word $w_l$ for our LF. We treat the context-only visual feature $\mathrm{c}^\text{L}$ as K, the word-focused visual feature $\mathrm{d}^\text{L}_l$ as Q, and the cross-modal feature $\mathrm{s}^\text{L}$ as V to perform LF through MHA. We also utilize the residual connection \cite{he2017resnet,vaswani2017att-all} in LF:
\begin{equation} \small
    \text{LF}(v^\text{L}_i) = v^\text{L}_i \oplus \text{MHA}(\mathrm{c}^\text{L}, \mathrm{d}^\text{L}, \mathrm{s}^\text{L}),
\end{equation}
where
\begin{equation} \small \notag
\begin{split}
    \mathrm{p}^\text{L} &= [v^\text{L}_i, P], \mathrm{q}^\text{L} = \{[v^\text{L}_i, P, e_{w_1}], ..., [v^\text{L}_i, P, e_{w_L}]\}, \\
    \mathrm{c}^\text{L} &= \text{Conv}^\text{L}(\mathrm{p}^\text{L}), \\
    \mathrm{d}^\text{L}_l &= \text{DotAtt}(\mathrm{p}^\text{L}, e_{w_l}) = \sum_{(h, w)} \text{softmax}(\mathrm{p}^\text{L} \cdot W^L_{\mathrm{d}} \cdot e_{w_l}^T)_{(h, w)} \cdot \mathrm{p}^\text{L}_{(h, w)}, \\
    \mathrm{s}^\text{L}_l &= \text{SelfAtt}(\mathrm{q}^\text{L}_l), \mathrm{s}^\text{L}_{l(h, w)} = \sum_{(x, y)} \text{softmax}(\mathrm{q}^\text{L}_l \cdot {\mathrm{q}^\text{L}_{l(h, w)}}^T)_{(x, y)} \cdot \mathrm{q}^\text{L}_{l(x, y)},
\end{split}
\end{equation}
and $W^L_{\mathrm{d}}$ is the learnable attention matrix between $\mathrm{p}^\text{L}$ and $e_w$. In this way, our LF fuses between visual context and word expression from SelfAtt and take the important portion of each token from DotAtt into consideration. 

For the global-level fusion (GF), it views the entire frame sequence $\{v_1, ..., v_N\}$ with the whole instruction $e_X$ to extract the global motion of the video. Similar to LF, we acquire the fused cross-modal feature $\mathrm{s}^\text{G}_n$ from SelfAtt, the context-only visual feature $\mathrm{c}^\text{G}_n$ from Conv$^\text{G}$, and the sentence-focused visual feature $\mathrm{d}^\text{G}_n$ from DotAtt for $v^\text{G}_n$. To model the entire video, we follow \cite{vaswani2017att-all}, where the video-level feature of $v_i$ can be represented as the relative weighted-sum over all frame-level $v$, and add on the positional encoding $\phi$ to incorporate the sequential order. We treat $\{\mathrm{s}^\text{G}_n\}$ as V, $\{\mathrm{c}^\text{G}_n\}$ as Q and $\{\mathrm{d}^\text{G}_n\}$ as K for the correlation between a frame pair, to perform GF through MHA:
\begin{equation} \small
    \text{GF}(v^\text{G}) = v^\text{G} \oplus \text{MHA}(\mathrm{c}^\text{G} \oplus \phi, \mathrm{d}^\text{G} \oplus \phi, \mathrm{s}^\text{G} \oplus \phi),
\end{equation}
where
\begin{equation} \small \notag
\begin{split}
    \mathrm{p}^\text{G} &= \{[v^\text{G}_1, P], ..., [v^\text{G}_N, P]\},~~ \mathrm{q}^\text{L} = \{[v^\text{G}_1, P, e_X], ..., [v^\text{G}_N, P, e_X]\}, \\
    \mathrm{c}^\text{G}_n &= \text{Conv}^\text{G}(\mathrm{p}^\text{G})_n,~ \mathrm{d}^\text{G}_n = \text{DotAtt}(\mathrm{p}^\text{G}_n, e_X),~ \mathrm{s}^\text{G}_n = \text{SelfAtt}(\mathrm{q}^\text{G}_n).
\end{split}
\end{equation}
By considering the correlation between frame with respect to the whole instruction from DotAtt, our GF models the video sequence as fused cross-modal feature from SelfAtt.

\vspace{0.5ex} \noindent\textbf{Encoder and Decoder.}
The encoder (Enc) in the multi-modal multi-level transformer $T$ serves to model the source video sequence $S$ with the given instruction $X$. Enc first adopts the local-level fusion (LF) to extract important portion from each single frame $v^s$ with each word embedding $e_w$; then the global-level fusion (GF) extracts the entire video motion with the sentence embedding $e_X$ as the cross-modal feature $f_i^s$:
\begin{equation} \small
    f_i^s = \text{GF}(\text{LF}(v^s, e_w), e_X)_i.
\end{equation}
During decoding, the decoder (Dec) also extracts the cross-modal feature $f^o_i$ as the same way from the previous generated frames $\{o_1, ..., o_{i-1}\}$. To acquire the decoding feature $d_i$ to generate the target frame, GF is first adopted to give the high-level concept of moving motion by the interaction between the cross-modal feature $f$ from source and target, where we treat $f^s$ as the fused feature (V). LF is applied for detailed specific property provided from word tokens $e_w$:
\begin{equation} \small
    f_i^o = \text{LF}(\text{GF}(\{v^o_1, ..., v^o_{i-1}\}, e_X | f^s)_i, e_w).
\end{equation}

In summary, the multi-modal multi-level transformer $T$ models the source video frame $v^s$ and the given instruction $\{e_X, e_w\}$, and considers previous generated target frames $\{o_1, ..., o_{i-1}\}$ to acquire the decoding feature $d_i$:
\begin{equation} \small
\label{eq:cmmlt}
    d_i = T(\{ o_1, ..., o_{i-1} \} | v^s, \{e_X, e_w\}).
\end{equation}

\subsection{Video Frame Generation}
With the decoding feature $d_i$ from $T$, we adopt ResBlocks \cite{miyato2018res-block} into the generator $U$ to scale up $d_i$ and synthesize into $\hat{o}_i$:
\begin{equation} \small
    \hat{o}_i = U(d_i), \qquad \hat{O} = \{\hat{o}_1, \hat{o}_2, ..., \hat{o}_N\}.
\end{equation}
We calculate the editing loss $\mathcal{L}_E$ by mean pixel difference using mean-square loss over each frame between $O$ and $\hat{O}$:
\begin{equation} \small
    \label{eq:mse}
    \mathcal{L}_E = \frac{1}{N} \sum_{i=1}^{N} \text{MSELoss}(o_i, \hat{o}_i).
\end{equation}

\vspace{0.5ex} \noindent\textbf{Dual Discriminator.}
Apart from the visual difference, we also consider the video quality of our generated $\hat{O}$. Similar to DVD-GAN \cite{clark2019dvd-gan}, we apply the dual discriminator $D$, where the frame discriminator $D_a$ improves the single frame quality and the temporal discriminator $D_t$ constrains the temporal consistency for a smooth output video $\hat{O}$. We treat $D_a$ as a binary classifier, which discriminates a target video frame $o$ is from ground-truth $O$ or our synthesized $\hat{O}$. Simultaneously, $D_t$ judges that if $K$ consecutive frames are smooth and consistent enough to be a real video fragment as the binary discrimination. The video quality loss $\mathcal{L}_{G}$ is computed for both frame quality and temporal consistency:
\begin{equation} \small
\begin{split}
    \label{eq:lg}
    \mathcal{L}_{\hat{a}} &= \frac{1}{N} \sum^{N}_{i=1} \log (1-D_a(\hat{o}_i)), \\
    \mathcal{L}_{\hat{t}} &= \frac{1}{M} \sum^{M}_{i=1} \log (1-D_t(\{\hat{o}_i, ..., \hat{o}_{i+K-1}\})), \\
    \mathcal{L}_G &= \mathcal{L}_{\hat{a}}+\mathcal{L}_{\hat{t}},
\end{split}
\end{equation}
where $M=N-K+1$. On the other hand, the dual discriminator $D$ is training to distinguish between $O$ and $\hat{O}$ by the following:
\begin{equation} \small
\begin{split}
    \label{eq:ld}
    \mathcal{L}_a &= \frac{1}{N} \sum^{N}_{i=1} ( \log (1-D_a(\hat{o}_i)) + \log (D_a(o_i)) ), \\
    \mathcal{L}_t &= \frac{1}{M} \sum^{M}_{i=1} ( \log (1-D_t(\{\hat{o}_i, ..., \hat{o}_{i+K-1}\})) \\
    &~~~~~~~~~~~~~~~~+ \log (D_t(\{o_i, ..., o_{i+K-1}\})) ), \\
    \mathcal{L}_D &= \mathcal{L}_a+\mathcal{L}_t.
\end{split}
\end{equation}
Therefore, they are optimized through an alternating minmax game:
\begin{equation} \small
    \min_{G} \max_{D} \mathcal{L}_G + \mathcal{L}_D.
\end{equation}

\begin{algorithm}[t]
\small
    \begin{algorithmic}[1]
        \State $T$: Multi-Modal Multi-Level Transformer
        \State $U$: Frame Generator
        \State $D$: Dual Discriminator, including $D_a$ and $D_t$
        \State $S$, $X$: Source Video, Instruction
        \State $O$: Ground-Truth Target Video
        \\
        \State Initialize $T$, $U$, $D$
        \While{TRAINING}
        \State $\{v_1, ..., v_N\}$ = 3D ResNet($S$)
        \State $e_X$, $\{e_{w_1}, ..., e_{w_N}\}$ = RoBERTa($X$)
        \For{$i \gets 1$ to $N$}
            \Comment{teacher-forcing training}
            \State $d_i$ $\gets$ $T$($\{ o_1, ..., o_{i-1} \} | v, \{e_X, e_w$\})
            \Comment{Eq.~\ref{eq:cmmlt}}
            \State $\hat{o}_i$ $\gets$ $U$($d_i$)
            \State $\mathcal{L}_E$ $\gets$ visual difference loss with $O$
            \Comment{Eq.~\ref{eq:mse}}
            \State $\mathcal{L}_G$ $\gets$ video quality loss from $D$
            \Comment{Eq.~\ref{eq:lg}}
            \State Update $T$ and $U$ by minimizing $\mathcal{L}_G$+$\mathcal{L}_E$
            \State $\mathcal{L}_D$ $\gets$ discrimination loss for $D$
            \Comment{Eq.~\ref{eq:ld}}
            \State Update $D$ by maximizing $\mathcal{L}_D$
        \EndFor
        \EndWhile
    \end{algorithmic}
    
    \caption{Multi-Modal Multi-level Transformer (M$^3$L)}
    \label{algo:learning}
\end{algorithm}

\subsection{Learning of M$^3$L}
Algo.~\ref{algo:learning} presents the learning process of the proposed multi-modal multi-level transformer (M$^3$L) for LBVE. Since LBVE is also a sequential generation process, we apply the widely used teacher-forcing training trick, where we feed in the ground-truth target frame $o_{i-1}$ instead of the predicted $\hat{o}_{i-1}$ from the previous timestamp to make the training more robust. We adopt the multi-modal multi-level transformer $T$ to model the source video and input instruction, and the frame generator $U$ to generate the target video frame. During training, we minimize the video quality loss $\mathcal{L}_G$ with the visual difference $\mathcal{L}_E$ to optimize M$^3$L. We also update the dual discriminator $D$, including the frame discriminator $D_a$ and the temporal discriminator $D_t$, by maximizing $\mathcal{L}_D$. Therefore, the entire optimization object can be summarized as:
\begin{equation} \small
    \min_{G, E} \max_{D} \mathcal{L}_G + \mathcal{L}_E + \mathcal{L}_D.
\end{equation}

\section{Datasets}

\begin{figure*}[t]
\centering
    \includegraphics[width=\linewidth]{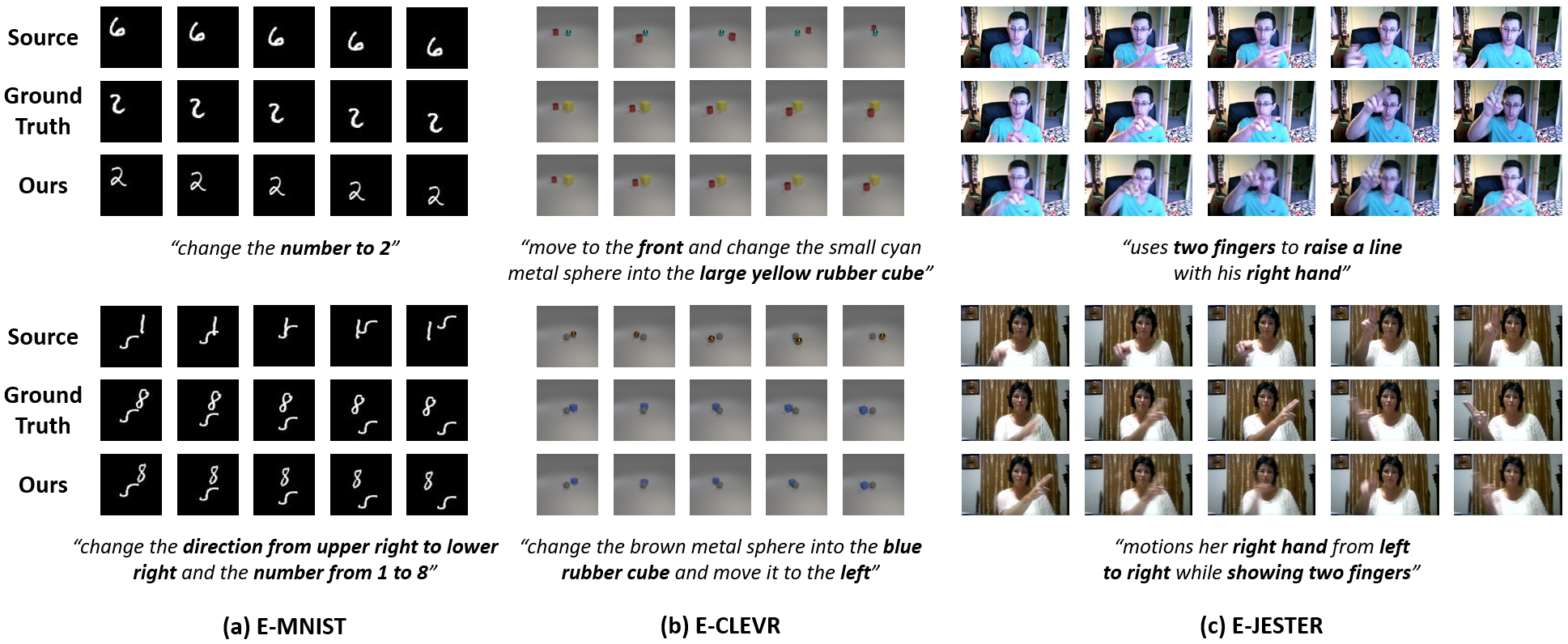}
    \vspace{-3ex}
    \caption{The sampled source videos, the ground-truth target videos, and the generated LBVE videos on all three datasets.}
    \vspace{-2ex}
    \label{fig:qual}
\end{figure*}

To the best of our knowledge, there is no dataset that supports video editing with the guided text. Therefore, we build three new datasets specially designed for LBVE, including two diagnostic datasets (E-MNIST and E-CLEVR) and one human gesture dataset (E-JESTER) for the language-based video editing (LBVE) task. An overview of our built datasets is shown in Table~\ref{table:dataset}, and examples of these three datasets are illustrated in Fig.~\ref{fig:qual}.

\vspace{0.5ex}
\noindent\textbf{E-MNIST.}
Extended from Moving MNIST \cite{lecun2010mnist,srivastava2015m-mnist}, the new E-MNIST dataset contains the instruction to describes the difference between two video clips. Hand-written numbers are moving along a specific direction and will reverse its direction if bumps into a boundary. The instructions include two kinds of editing actions: \emph{content replacing} is to replace the specific number with the given one, and \emph{semantic manipulation} changes the starting direction for different moving motion. We prepare two levels of E-MNIST, S-MNIST and D-MNIST. S-MNIST is an easier one and includes only a single number, so the model only needs to replace the number or change the moving direction at a time. There are two numbers in the advanced D-MNIST, where the model is required to perceive which number should be replaced and which starting direction should be changed simultaneously. For both S-MNIST and D-MNIST, there are 11,808 pairs of source-target video. 

\begin{table}[t]
\scriptsize \centering
    \begin{tabular}{crrrrc}
        \toprule
        \textbf{Dataset} & \textbf{\#Train} & \textbf{\#Test} & \textbf{\#Frame} & \textbf{\#Word} & \textbf{Resolution} \\
        \midrule
        S-MNIST & 11,070 & 738 & 354,240 & 5.5 & 64x64 \\
        D-MNIST & 11,070 & 738 & 354,240 & 16.0 & 64x64 \\
        E-CLEVR & 10,133 & 729 & 21,7240 & 13.4 & 128x128 \\
        E-JESTER & 14,022 & 885 & 59,508 & 9.9 & 100x176 \\
        \bottomrule
    \end{tabular}
    \vspace{-1ex}
    \caption{The statistics of our collected datsets.}
    \label{table:dataset}
    \vspace{-4ex}
\end{table}

\vspace{0.5ex}
\noindent\textbf{E-CLEVR.}
Following CATER \cite{girdhar2020cater}, we create each frame and combine them as the video in our E-CLEVR upon the original CLEVR dataset \cite{johnson2017clevr}. Each example consists of a pair of source-target videos with an instruction described the semantic altering. The editing action includes changing the property of the specific object and placing the moving object into a particular given final position. E-CLEVR contains plentiful object properties (e.g., color, shape, size, ...) and different relative positions of the final target. To highlight the importance of visual perception, not all aspects of the property will change; only the mentioned properties like the color and shape should be changed but keeps others the same. We generate 10,862 examples for E-CLEVR.

\begin{table*}[t]
\scriptsize \centering
    \begin{tabular}{lcccccccccccccc}
        \toprule
        ~ & \multicolumn{3}{c}{\textbf{S-MNIST}} & ~ & \multicolumn{3}{c}{\textbf{D-MNIST}} & ~ & \multicolumn{3}{c}{\textbf{E-CLEVR}} & ~ & \multicolumn{2}{c}{\textbf{E-JESTER}} \\
        \cmidrule{2-4} \cmidrule{6-8} \cmidrule{10-12} \cmidrule{14-15}
        ~ & VAD $\downarrow$ & OA $\uparrow$ & mIoU $\uparrow$ & ~ & VAD $\downarrow$ & OA $\uparrow$ & mIoU $\uparrow$ & ~ & VAD $\downarrow$ & OA $\uparrow$ & mIoU $\uparrow$ & ~ & VAD $\downarrow$ & GA $\uparrow$ \\
        \midrule
        pix2pix \cite{isola2017pix2pix} & 2.06 & 96.6 & 74.3 & ~ & 3.05 & 87.7 & 64.1 & ~ & 2.84 & 80.4 & 60.5 & ~ & 2.00 & 8.6 \\
        vid2vid \cite{wang2018vid2vid} & 1.30 & 97.0 & 88.6 & ~ & 2.30 & 87.5 & 77.9 & ~ & 2.21 & 80.5 & 69.3 & ~ & 1.62 & 82.0 \\
        E3D-LSTM \cite{wang2019e3d-lstm} & 1.29 & 97.8 & 92.8 & ~ & 2.10 & 90.4 & 81.3 & ~ & 2.11 & 83.1 & 72.2 & ~ & 1.55 & 83.6 \\
        M$^3$L (Ours) & \textbf{1.28} & \textbf{99.7} & \textbf{93.6} & ~ & \textbf{1.90} & \textbf{93.2} & \textbf{84.7} & ~ & \textbf{1.96} & \textbf{84.5} & \textbf{78.4} & ~ & \textbf{1.44} & \textbf{89.3} \\
        \bottomrule
    \end{tabular}
    \vspace{-1ex}
    \caption{The overall testing results of the baselines and our M$^3$L under the E-MMIST, E-CLEVR, and E-JESTER datasets.}
    \label{table:quan}
    \vspace{-2ex}
\end{table*}

\vspace{0.5ex}
\noindent\textbf{E-JESTER.}
Toward human action understanding, 20BN-JESTER \cite{materzynska2019jester} builds a large gesture recognition dataset. Each actor performs different kinds of gesture moving in front of the camera, which brings out 27 classes in total. This setting is appropriate to the video editing task where the source-target videos are under the same scenario (same person in the same environment) but with different semantics (different hand gestures). To support our LBVE task, we prepare pairs of clips from the same person as the source-target videos and collect the human-labeled instruction by Amazon Mechanical Turk (AMT)\footnote{Amazon Mechanical Turk: \url{https://www.mturk.com/}}. A person can exist in both training and testing sets but with different gestures. We ensure that there is no overlapping of the same person-gesture pairs between train/test splits. In this way, we can have the natural video whose scenario is preserved, but semantic is changing with natural guided text for our E-JESTER dataset, which can be a sufficient first step for LBVE. There are 14,907 pairs in E-JESTER.

\begin{table}[t]
\scriptsize \centering
    \begin{tabular}{cccc}
        \toprule
        ~ & ~ & \multicolumn{2}{c}{\textbf{E-JESTER}} \\
        \cmidrule{3-4}
        Instruction & MLF & VAD $\downarrow$ & GA $\uparrow$ \\
        \midrule
        \ding{55} & \ding{55} & 1.99 & 4.7 \\
        \ding{51} & \ding{55} & 1.50 & 85.4 \\
        \ding{51} & \ding{51} & \textbf{1.44} & \textbf{89.3} \\
        \bottomrule
    \end{tabular}
    \vspace{-1ex}
    \caption{The ablation results when without the instruction or MLF.}
    \label{table:abl}
    \vspace{-3ex}
\end{table}

\section{Experiments}
\subsection{Experimental Setup}
\vspace{0.5ex} \noindent\textbf{Evaluation Metrics.}
\begin{itemize}[topsep=0pt, noitemsep, leftmargin=*]
    \item \textbf{VAD}: Inspired by IS \cite{salimans2016is} and FID \cite{hausel2017fid}, we apply 3D CNN and compute the video activation distance (VAD) as the mean L2 distance between video feature. Specifically, ResNeXt \cite{xie2017resnext} is adopted for the diagnostic E-MNIST and E-CLEVR dataset. Besides, we utilize I3D \cite{carreira2017i3d} to extract the action video feature for E-JESTER. A lower VAD means that videos are more related to each other.
    \item \textbf{OA}: Apart from the visual-base evaluation, we consider the object accuracy (OA) for E-MNIST and E-CLEVR. OA is calculated by the correctness of the presented objects in the target video from a pre-trained object detector\footnote{\label{object-detector} We have more than 99\% OA and 95\% mIoU of our pre-trained object detector, which can precisely evaluate E-MNIST and E-CLEVR.}. A higher OA shows that the model can edit specific properties of the mentioned object from the instruction.
    \item \textbf{mIoU}: We also evaluate the position of objects for E-MNIST and E-CLEVR via mean Intersection over Union (mIoU) between generated and ground-truth results. mIoU is averaged from each frame in the video also based on the pre-trained object detector. A higher mIoU indicates that the model is able to manipulate the object into the mentioned relative position.
    \item \textbf{GA}: We report the gesture accuracy (GA) for E-JESTER, which is calculated as the gesture classification accuracy of the edited video by MFFs\footnote{MFFs (\url{https://github.com/okankop/MFF-pytorch}) has 96\% accuracy on JESTER and serves for evaluating E-JESTER.}. Although the generated video may not be the same as the ground truth, a higher GA represents that the model is able to follow the guided text and generate the corresponding type of gesture.
\end{itemize}

\vspace{0.5ex}
\noindent\textbf{Baselines.}
Since our LBVE is a brand new task, there is no existing baselines. We consider following methods conditioning on an instruction, by concatenating the languistic feature, to carry out LBVE as the compared baselines.
\begin{itemize}[topsep=0pt, noitemsep, leftmargin=*]
    \item \textbf{pix2pix} \cite{isola2017pix2pix}: pix2pix is an image-to-image translation approach. For the sake of video synthesis, we process the source video frame-by-frame to perform pix2pix.
    \item \textbf{vid2vid} \cite{wang2018vid2vid}: vid2vid applies the temporal discriminator for better video-to-video synthesis, which considers several previous frames to model the translation.
    \item \textbf{E3D-LSTM} \cite{wang2019e3d-lstm}: E3D-LSTM incorporates 3D CNN into LSTM for video prediction. We treat the source video as the given video and predict the remaining part as the target video.
\end{itemize}

\vspace{0.5ex}
\noindent\textbf{Implementation Detail.}
We apply 3-layer ResBlocks \cite{miyato2018res-block} into the 3D ResNet and the generator $U$ with kernel size 3 and stride 1 in the first layer. In particular, we incorporate 1-layer self-attention for better frame generation into $U$ following SAGAN \cite{zhang2019sa-gan}. The visual feature dimension $C_v$ is 256 and the language feature dimension $C_x$ is 1024 from RoBERTa \cite{liu2019roberta}. Adam \cite{kingma2015adam} is adopted to optimize through our multi-modal multi-level transformer (M$^3$L) with learning rate 3e-4 for the visual difference loss $\mathcal{L}_E$, and learning rate 1e-4 for $\mathcal{L}_G$ and $\mathcal{L}_D$ from the dual discriminator $D$. 

\begin{table}[t]
\scriptsize \centering
    \begin{tabular}{cccccccc}
        \toprule
        ~ & \multicolumn{3}{c}{\textbf{D-MNIST}} & ~ & \multicolumn{3}{c}{\textbf{E-CLEVR}} \\
        \cmidrule{2-4} \cmidrule{6-8}
        MLF & VAD $\downarrow$ & OA $\uparrow$ & mIoU $\uparrow$ & ~ & VAD $\downarrow$ & OA $\uparrow$ & mIoU $\uparrow$ \\
        \midrule
        \ding{55} & 2.64 & 82.6 & 73.6 & ~ & 2.32 & 70.1 & 66.6 \\
        \ding{51} & \textbf{2.35} & \textbf{87.5} & \textbf{79.1} & ~ & \textbf{2.29} & \textbf{76.7} & \textbf{71.5} \\
        \bottomrule
    \end{tabular}
    \vspace{-1ex}
    \caption{Zero-shot generalization under D-MNIST and E-CLEVR.}
    \label{table:zs}
    \vspace{-3ex}
\end{table}

\subsection{Quantitative Results}
Table~\ref{table:quan} shows the overall testing results compared between the baselines and ours M$^3$L. pix2pix only adopts image-to-image translation, resulting in insufficient output video (\textit{e.g.,} 64.1 mIoU under D-MNIST and 2.84 VAD under E-CLEVR). Even if vid2vid and E3D-LSTM consider temporal consistency, the lack of explicit cross-modal fusion still makes them difficult to perform LBVE. While, our M$^3$L, which incorporates the multi-level fusion (MLF), can fuse between vision-and-language with different levels and surpass all baselines. In particular, M$^3$L achieves the best results across all metrics under all diagnostic datasets (\textit{e.g.,} 99.7 OA under S-MNIST, 84.7 mIoU under D-MNIST, and 1.96 VAD under E-CLEVR).

Similar trends can be found on the natural E-JESTER dataset. pix2pix only has 8.6\% GA, which shows that it cannot produce a video with the correct target gesture. Although vid2vid and E3D-LSTM may have similar visual measurement scores to our approach, M$^3$L achieves the highest 89.3\% GA. The significant improvement of GA demonstrates that the proposed MLF benefits not only the visual quality but also the semantic of the predicted video and makes it more corresponding to the given instruction.

\subsection{Ablation Study}
\vspace{0.5ex} \noindent\textbf{Ablation Results.}
Table~\ref{table:abl} presents the testing results of the ablation setting under E-JESTER. If without the given instruction, the model lacks the specific editing target and results in poor 1.99 VAD and 4.7\% GA. The performance comprehensively improves when incorporating our proposed multi-level fusion (MLF) (\textit{e.g.,} VAD from 1.50 down to 1.44 and GA from 85.4\% up to 89.3\%). The multi-level modeling from MLF benefits not only the understanding between video and instruction, but also leads to accurate frame generation. The above ablation results show that the instruction is essential under the video editing task, and our MLF further helps to perform LBVE.

\begin{table}[t]
\scriptsize \centering
    \begin{tabular}{lccc}
        \toprule
        ~ & \textbf{w/ MLF} & \textbf{w/o MLF} & \textbf{Tie} \\
        \midrule
        Video Quality & \textbf{67.1\%} & 27.1\% & 5.8\% \\
        Video-Instruction Alignment & \textbf{53.3\%} & 35.1\% & 11.6\% \\
        Siml. to GT Video & \textbf{59.6\%} & 28.9\% & 11.6\% \\
        \bottomrule
    \end{tabular}
    \vspace{-1ex}
    \caption{Human evaluation on E-JESTER with aspects of video quality, video-instruction alignment, and similarity to GT video.}
    \label{table:human}
    \vspace{-4ex}
\end{table}

\vspace{0.5ex} \noindent\textbf{Zero-Shot Generalization.}
To further investigate the generalizability of M$^3$L, we conduct a zero-shot experiment for both the D-MNIST and E-CLEVR datasets. In D-MNIST, there are 40 different object-semantic combinations\footnote{\textbf{D-MNIST}: 10 different numbers and 4 different directions}. We remove out 10 of them in the training set (\textit{e.g.,} number 1 with upper left or number 3 with lower down) and evaluate under the complete testing set. For E-CLEVR, we filter out 12 kinds (\textit{e.g.,} small gray metal sphere or large purple rubber cube) from the total 96 possibilities\footnote{\textbf{E-CLEVR}: 3 shapes, 8 colors, 2 materials, and 2 shapes}. This testing scenario is widely used to evaluate new combinations of object-semantic pairs that are not seen during training~\cite{fu2020sscr,girdhar2020cater,chaplot2018ga}. The results are shown in Table~\ref{table:zs}. Due to the lack of object properties or moving semantics, the model has a significant performance drop under the zero-shot settings. While, our proposed MLF helps the property and moving motion for both video perception and generation by multi-modal multi-level fusion. Therefore, MLF still improves the generalizability (\textit{e.g.,} OA from 82.6 up to 87.5 under D-MNIST and mIoU from 66.6 up to 71.5 under E-CLEVR) even if training with the zero-shot examples.

\vspace{0.5ex} \noindent\textbf{Inference Efficiency.}
As a video processing task, not only the performance but also the efficiency is important of the editing framework. When using only the CPU, it carries out the E-JSTER with about 11.9 FPS, where the processed frame is 128x128. With the acceleration from the GPU (TITAN X), the model can further achieve 35.8 FPS, which is faster than the real-time requirement (24 FPS). The results show that our M$^3$L with the multi-level fusion (MLF) can carry out the LBVE task for practical usage efficiently.

\vspace{0.5ex} \noindent\textbf{Human Evaluation.}
Apart from the quantitative results, we also investigate the quality of the generated video from the human aspect. Table~\ref{table:human} demonstrates the comparison between without and with MLF. We randomly sample 75 examples and ask three following questions: (1) Which video has better quality; (2) Which video corresponds more to the given instruction; (3) Which video is more similar to the ground-truth target video. Each example is assigned to 3 different MTurkers to avoid evaluation bias. Firstly, about 67\% think that generated videos from MLF have better quality. Moreover, more than 50\% of Mturkers denote that the target videos produced from MLF correspond more to the instruction and are also more similar to the ground truth. The results of the human evaluation indicate that our MLF not only helps improve the generating quality but also makes the target video more related to the guided text. 

\vspace{0.5ex} \noindent\textbf{Qualitative Results.}
Fig.~\ref{fig:qual} shows the keyframes of the generated examples of LBVE on all three datasets. For E-MNIST, we have to recognize which number should be replaced and which one will change the moving semantic. Note that the instruction only tells the replacing number, but without the style, thus our model replaces with another kind of number 2 under S-MNIST. Under the advanced D-MNIST dataset, our model can replace with the number 8 and move the number 5 along the lower right with multi-level fusion. The challenge of E-CLEVR is to transform object properties and move to the different target positions related to the fixed object. The visualization examples show that our model can understand the linguistic to change the specific object into the correct properties. Also, it has the spatial concept that can perceive the final related position and maintain the moving motion. The E-JESTER dataset, which contains nature video and human-labeled instruction, requires the link of the complex natural language with the human gesture action. The presented video indicates that our model can not only preserve a similar scenario (the background and the person) but also generate the visual motion of the corresponding gesture.

\section{Conclusion}
We introduce language-based video editing (LBVE), a novel task that allows the model to edit, guided by a natural text, a source video into a target video. We present multi-modal multi-level transformer (M$^3$L) to dynamically fuse video perception and language understanding at multiple levels. For the evaluation, we release three new datasets containing two diagnostic and one natural video with human-labeled text. Experimental results show that our M$^3$L is adequate for video editing, and LBVE can bring out a new field toward vision-and-language research.

\vspace{0.5ex} \noindent\textbf{Acknowledgments.} Research was sponsored by the U.S. Army Research Office and was accomplished under Contract Number W911NF-19-D-0001 for the Institute for Collaborative Biotechnologies. The views and conclusions contained in this document are those of the authors and should not be interpreted as representing the official policies, either expressed or implied, of the U.S. Government. The U.S. Government is authorized to reproduce and distribute reprints for Government purposes notwithstanding any copyright notation herein.

\clearpage

\onecolumn \appendix

\begin{figure*}[!h]
    \centering
    \includegraphics[width=\linewidth]{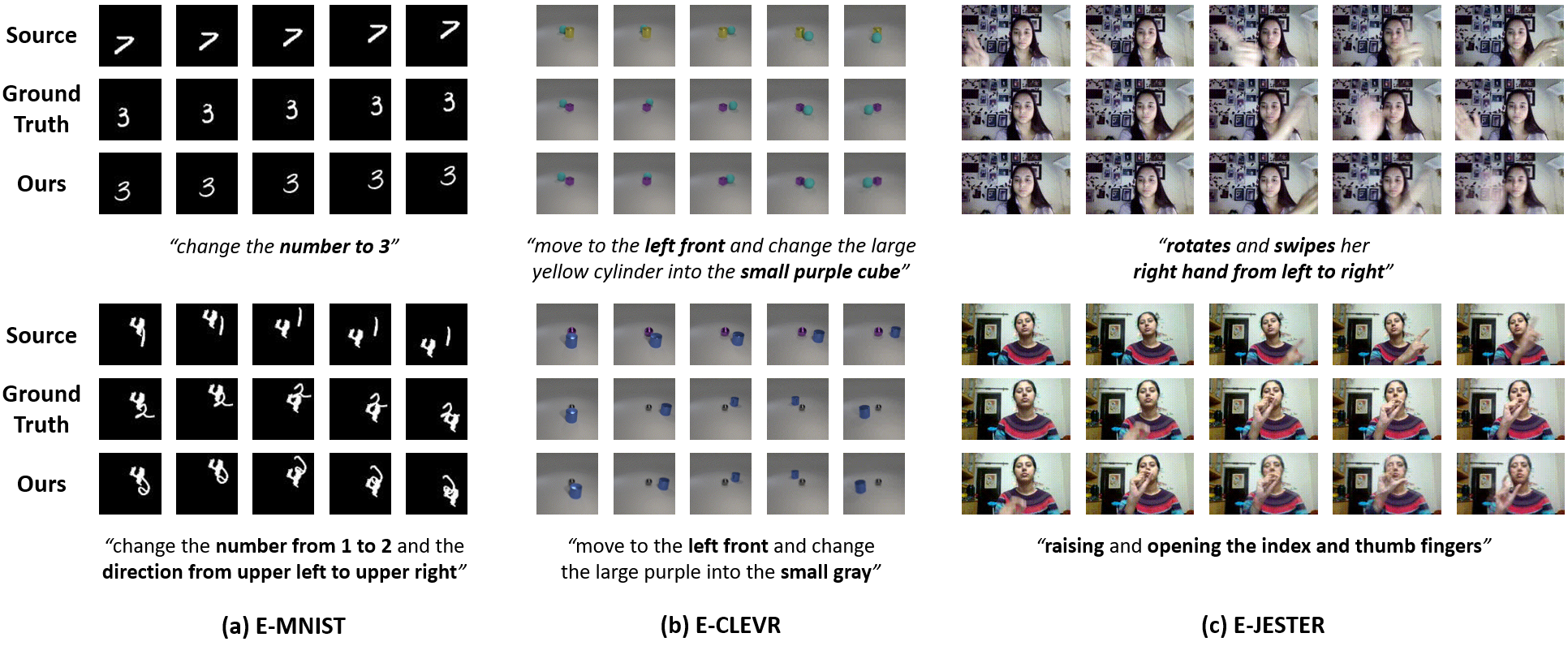}
    \vspace{-3ex}
    \captionof{figure}{The sampled source videos, the ground-truth target videos, and the generated LBVE videos on all three datasets.}
    \vspace{-1ex}
\end{figure*}

\section{Zero-shot Generalization under E-JESTER}
We conduct the zero-shot setting on E-JESTER, where the people in the testing set do not exist during training. We evaluate the generalizability of a model through editing an unseen person with a specific gesture. The results are summarized in Table~\ref{table:zs-jester}. pix2pix~\cite{isola2017pix2pix}, which only treats single frame translation, performs the worst. Both vid2vid~\cite{wang2018vid2vid} and E3D-LSTM~\cite{wang2019e3d-lstm} result in a significant performance drop under the zero-shot setting (\textit{e.g.,} vid2vid drops from 82.0 GA to 73.8 and E3D-LSTM ups from 1.55 VAD to 1.79). In contrast, with the multi-level fusion (MLF) over different levels of video-and-language reasoning, our M$^\text{3}$L still maintains the lowest 1.51 VAD and the highest 86.0 GA, even encountering an unseen person.
\begin{table}[!ht]
\scriptsize \centering
    \begin{tabular}{lcrccr}
        \toprule
        ~ & \multicolumn{2}{c}{\textbf{E-JESTER} (Full)} & ~ & \multicolumn{2}{c}{\textbf{E-JESTER} (Zero-shot)} \\
        \cmidrule{2-3} \cmidrule{5-6} ~ & VAD $\downarrow$ & GA $\uparrow$ & ~ & VAD $\downarrow$ & GA $\uparrow$ \\
        \midrule
        pix2pix~\cite{isola2017pix2pix} & 2.00 & 8.6 & ~ & 2.42 & 8.7 \\
        vid2vid~\cite{wang2018vid2vid} & 1.62 & 82.0 & ~ & 1.84 & 73.8 \\
        E3D-LSTM~\cite{wang2019e3d-lstm} & 1.55 & 83.6 & ~ & 1.79 & 78.4 \\
        M$^\text{3}$L (Ours) & \textbf{1.44} & \textbf{89.3} & ~ & \textbf{1.51} & \textbf{86.0} \\
        \bottomrule
    \end{tabular}
    \vspace{-1ex}
    \caption{Zero-shot Generalization under E-JESTER.}
    \label{table:zs-jester}
    \vspace{-2ex}
\end{table}

\section{Human Evaluation of Baselines}
We conduct a human evaluation with 30 E-JESTER examples over all baselines. Table~\ref{table:suppl-human} shows the mean ranking score (from 1 to 4, the higher is better) under different aspects. In general, videos produced by our M$^3$L have higher quality. Furthermore, the proposed MLF makes the editing result more related to the guided text.
\begin{table}[!ht]
\scriptsize \centering
    \begin{tabular}{lcccc}
        \toprule
        ~ & pix2pix & vid2vid & E3D-LSTM & M$^3$L \\
        \midrule
        Video Quality & 2.07 & 2.47 & 2.50 & \textbf{2.97} \\
        Video-Instruction Alignment & 1.67 & 2.27 & 2.37 & \textbf{3.67} \\
        Similarity to GT Video & 1.60 & 2.40 & 2.63 & \textbf{3.37} \\
        \bottomrule
    \end{tabular}
    \vspace{-1ex}
    \caption{Human evaluation (mean ranking score from 1 to 4, the higher is better) on E-JESTER.}
    \label{table:suppl-human}
    \vspace{-2ex}
\end{table}

\section{Ablation of MLF/Discriminator}
Table \ref{table:suppl-abl} illustrates the ablation study of multi-level fusion (MLF), including local-level (LF) and global-level fusion (GF), and dual discriminator (Dual-D) on E-CLEVR. Comparing row (b) and (c) with (a), LF contains better local perception (higher OA) between object properties and word tokens, and GF benefits the global motion (lower VAD and higher mIoU). Row (d) further shows that combining LF and GF as MLF can help both. In the end (row (e)), Dual-D enhances the video quality, leading to a comprehensive improvement.
\begin{table}[!ht]
\scriptsize \centering
    \begin{tabular}{ccccccc}
        \toprule
        ~ & LF & GF & Dual-D & VAD $\downarrow$ & OA $\uparrow$  & mIoU $\uparrow$ \\
        \midrule
        (a) & \ding{55} & \ding{55} & \ding{55} & 2.19 & 82.4 & 70.5 \\
        (b) & \ding{51} & \ding{55} & \ding{55} & 2.25 & 83.4 & 71.7 \\
        (c) & \ding{55} & \ding{51} & \ding{55} & 2.04 & 83.1 & 74.6 \\
        (d) & \ding{51} & \ding{51} & \ding{55} & \underline{2.02} & \underline{83.6} & \underline{75.3} \\
        (e) & \ding{51} & \ding{51} & \ding{51} & \textbf{1.96} & \textbf{84.5} & \textbf{78.4} \\
        \bottomrule
    \end{tabular}
    \vspace{-1ex}
    \caption{Ablation study of MLF/Discriminator on E-CLEVR.}
    \label{table:suppl-abl}
    \vspace{-2ex}
\end{table}

\section{Multi-Modal Baseline}
We consider GeNeVA~\cite{elnouby2018ilbie}, iterative-base LBIE, as the multi-modal baseline. For each turn, we feed in the instruction and generate a frame based on previous results and the encoded source video from LSTM. Then we compose all iterative frames as the editing video. Table~\ref{table:geneva} shows the evaluation on E-CLVER. GeNeVA has better OA and MIoU than E3D-LSTM by the self-attention module over the visual-and-linguistic feature. Upon cross-modal attention, M$^3$L further considers multi-level fusion (MLF), leading to the best results on all metrics. 
\begin{table}[!ht]
\scriptsize \centering
    \begin{tabular}{lccc}
        \toprule
        Method & VAD $\downarrow$ & OA $\uparrow$ & mIoU $\uparrow$ \\
        \midrule
        E3D-LSTM & \underline{2.11} & 83.1 & 72.2 \\
        GeNeVA & 2.13 & \underline{83.3} & \underline{74.5} \\
        M$^3$L & \textbf{1.96} & \textbf{84.5} & \textbf{78.4} \\
        \bottomrule
    \end{tabular}
    \vspace{-1ex}
    \caption{The testing results of GeNeVA on E-CLEVR.}
    \label{table:geneva}
    \vspace{-2ex}
\end{table}

\section{Limitation and Social Impact}
Our M$^3$L framework treats source/target videos as fully-supervised training, which may fail for out-domain scenes and instructions. We can exploit pretrained visual-linguistic alignment (\textit{e.g.,} CLIP~\cite{shi2020clip}) to boost the editing result weakly-supervisedly. Besides, there may be an authenticity doubt for those edited videos. To mitigate this issue, we train a binary video classifier, which achieves 93\% real/fake accuracy on E-JESTER. It shows that such video forensics can help video authentication of the potential negative impact.

{\small
\bibliographystyle{ieee_fullname}
\bibliography{egbib}
}

\end{document}